\documentclass{article}

 \usepackage[main, final]{neurips_2025}

\makeatletter
\renewcommand{\@noticestring}{} 
\makeatother

\usepackage[utf8]{inputenc} 
\usepackage[T1]{fontenc}    
\usepackage{hyperref}       
\usepackage{url}            
\usepackage{booktabs}       
\usepackage{amsfonts}       
\usepackage{nicefrac}       
\usepackage{microtype}      
\usepackage{xcolor}         
\usepackage{graphicx}
\usepackage{amsmath} 

\title{Frequency Regularization: Unveiling the Spectral Inductive Bias of Deep Neural Networks}

\author{
  Jiahao Lu \\
  School of Artificial Intelligence\\
  The Chinese University of Hong Kong, Shenzhen\\
  \texttt{lujiahaohk@163.com} \\}

\begin{document}

\maketitle

\begin{abstract}
  Regularization techniques such as L2 regularization (Weight Decay) and Dropout are fundamental to training deep neural networks, yet their underlying physical mechanisms regarding feature frequency selection remain poorly understood. In this work, we investigate the \textit{Spectral Bias} of modern Convolutional Neural Networks (CNNs). We introduce a \textbf{Visual Diagnostic Framework} to track the dynamic evolution of weight frequencies during training and propose a novel metric, the \textbf{Spectral Suppression Ratio (SSR)}, to quantify the "low-pass filtering" intensity of different regularizers. By addressing the aliasing issue in small kernels (e.g., $3\times3$) through discrete radial profiling, our empirical results on ResNet-18 and CIFAR-10 demonstrate that L2 regularization suppresses high-frequency energy accumulation by over \textbf{3$\times$} compared to unregularized baselines. Furthermore, we reveal a critical \textbf{Accuracy-Robustness Trade-off}: while L2 models are sensitive to broadband Gaussian noise due to over-specialization in low frequencies, they exhibit superior robustness against \textbf{high-frequency information loss} (e.g., low resolution), outperforming baselines by $>6\%$ in blurred scenarios. This work provides a signal-processing perspective on generalization, confirming that regularization enforces a strong spectral inductive bias towards low-frequency structures.
Code is available at \url{https://github.com/lujiahao760/FrequencyRegularization}.

\end{abstract}

\section{Introduction}
\label{sec:intro}

Deep neural networks have achieved remarkable success across diverse domains, yet a fundamental question persists: how do these over-parameterized models generalize effectively to unseen data? Standard regularization techniques—most notably \textbf{L2 regularization} (Weight Decay) and \textbf{Dropout}—are ubiquitous in training modern architectures like ResNet \citep{he2016deep} to mitigate overfitting. The conventional interpretation frames L2 as a complexity constraint: by penalizing weight magnitudes, it restricts the model's hypothesis space from a VC-dimension perspective. However, this view fails to address two critical gaps: it does not clarify \textit{which specific features} the model prioritizes under such constraints, nor does it describe the physical evolution of learned representations during training.

Recent theoretical advances have proposed the "Spectral Bias" or "Frequency Principle" \citep{rahaman2019spectral, xu2019frequency}, positing that neural networks inherently learn low-frequency components of target functions first, before fitting high-frequency details. While these theories offer valuable insights, they are largely derived from simple Multi-Layer Perceptrons (MLPs) or synthetic datasets. Crucially, there is a lack of practical diagnostic tools to visualize and quantify this spectral behavior in modern CNNs—which rely heavily on $3\times3$ small kernels—trained on real-world image datasets. Moreover, the specific impact of explicit regularization strategies (e.g., L2, Dropout) on the \textit{dynamics} of spectral learning remains empirically underexplored.

In this work, we bridge the gap between theoretical spectral bias and practical deep learning engineering by treating neural network training as a signal processing problem. Convolutional kernels, we argue, act as frequency-domain filters: their weights' spectral properties directly determine which frequency components of input data the model prioritizes. By analyzing the 2D Discrete Fourier Transform (DFT) of weights in ResNet-18 trained on CIFAR-10, we provide both visual and quantitative evidence that regularization—particularly L2—functions as an explicit \textbf{Low-Pass Filter}.

Our contributions are structured into a cohesive diagnostic framework:
\begin{enumerate}
    \item \textbf{Visual Diagnostic Framework for Spectral Dynamics:} We propose a signal-processing lens to analyze regularization's inductive bias. Unlike prior work focusing on static weight spectra, our framework tracks the \textit{dynamic evolution} of frequency components throughout training, revealing how L2 regularization steers the optimization trajectory toward low-frequency solutions.
    
    \item \textbf{Small-Kernel Spectral Analysis:} Standard spectral methods fail for $3\times3$ kernels (ubiquitous in modern CNNs) due to severe aliasing and grid quantization errors. We address this limitation with a \textbf{Discrete Radial Profiling} algorithm, enabling precise frequency quantification without information loss.
    
    \item \textbf{SSR Metric and Robustness Trade-off:} We introduce the \textbf{Spectral Suppression Ratio (SSR)}, a scale-invariant metric to quantify low-pass filtering intensity. Using SSR, we uncover a critical trade-off: L2-induced low-frequency specialization enhances robustness to high-frequency information loss (e.g., blur, low resolution) but increases sensitivity to broadband noise.
\end{enumerate}

\section{Related Work}
\label{sec:related_work}

\paragraph{Regularization and Generalization.}
Understanding DNN generalization remains a core challenge in machine learning theory. Classical statistical learning (e.g., VC-dimension) cannot explain why over-parameterized networks avoid overfitting \citep{neyshabur2014search}. Explicit regularization techniques—including Weight Decay (L2) and Dropout \citep{srivastava2014dropout}—are standard remedies, but their implicit biases are not fully understood. Traditional interpretations frame L2 as a norm constraint on weight space, but this does not capture its impact on learned features. Our work advances this line of inquiry by providing a frequency-domain interpretation of regularization, linking it to selective suppression of high-frequency components.

\paragraph{Spectral Bias in Neural Networks.}
A growing body of work uses Fourier analysis to study neural network learning. \citet{rahaman2019spectral} and \citet{xu2019frequency} empirically and theoretically demonstrated that networks exhibit spectral bias: low-frequency components are learned faster than high-frequency ones. \citet{basri2020frequency} further explained this via the Neural Tangent Kernel (NTK) regime. However, these studies focus on theoretical bounds or simple MLPs on synthetic data. We extend this analysis to modern CNNs (ResNet-18) on natural images (CIFAR-10), validating spectral bias in practical settings and linking it to explicit regularization.

\paragraph{Frequency and Robustness.}
The relationship between frequency components and model robustness has gained significant attention. \citet{yin2019fourier} showed that CNNs are hypersensitive to high-frequency adversarial perturbations, while \citet{geirhos2019imagenet} argued that standard CNNs prioritize high-frequency "texture" features, whereas humans rely on low-frequency "shape." These works suggest that biasing models toward low frequencies could improve robustness. Our empirical results confirm this: L2 regularization shifts models from texture to shape bias, enhancing robustness to resolution degradation (high-frequency loss).

\paragraph{The Generalization Puzzle.}
Classical learning theory attributes generalization to model complexity constraints, but \citet{zhang2021understanding} showed that DNNs can memorize random labels, challenging this view. Research has since shifted to implicit regularization (e.g., gradient descent dynamics \citep{neyshabur2014search}) and explicit regularizers. However, how these techniques shape internal feature representations—particularly in terms of frequency—remains unclear. Our work fills this gap by demonstrating that regularization's generalization benefit stems from its spectral filtering effect.

\section{Method}
\label{sec:method}

To investigate regularization's underlying mechanisms, we propose a spectral analysis framework that treats CNN weights as 2D discrete signals. Below, we formalize our approach for extracting, transforming, and quantifying weight frequency components—with a focus on addressing the unique challenges of modern CNNs.

\subsection{Preliminaries: Spectral Representation of Weights}
Consider a standard CNN layer parameterized by a weight tensor $\mathcal{W} \in \mathbb{R}^{C_{out} \times C_{in} \times K \times K}$, where $K$ denotes the spatial kernel size. We focus analysis on individual 2D spatial kernels $W \in \mathbb{R}^{K \times K}$, as their frequency properties directly shape the layer's filtering behavior.

To characterize frequency components, we compute the 2D Discrete Fourier Transform (DFT) of each kernel $W$. The spectral representation $\mathcal{F}(W)$ at frequency coordinates $(u, v)$ is:
\begin{equation}
    \mathcal{F}(W)[u, v] = \sum_{x=0}^{K-1} \sum_{y=0}^{K-1} W[x, y] \cdot e^{-j2\pi\left(\frac{ux}{K} + \frac{vy}{K}\right)}
\end{equation}
We then calculate the \textbf{Power Spectral Density (PSD)} as $P(u, v) = |\mathcal{F}(W)[u, v]|^2$, which quantifies the energy of each frequency component. Following standard practice, we apply a spectral shift to center the zero-frequency (DC) component at the spectrum's origin $(0,0)$, enabling intuitive visualization: low frequencies cluster at the center, and high frequencies at the periphery.

\subsection{Algorithm: Discrete Radial Profiling for Small Kernels}
\label{sec:method_drp}

\textbf{Challenge of Small Kernels:} Standard spectral analysis condenses 2D spectra into 1D profiles via radial averaging—grouping frequencies into concentric annular bins by radius $r$. However, modern CNNs (e.g., ResNet) rely almost exclusively on $3\times3$ kernels. For such low-resolution grids, continuous binning causes severe aliasing, empty bins, and statistical instability—rendering standard tools ineffective.

\textbf{Proposed Solution: Discrete Radial Profiling.} We design an algorithm tailored to microscopic kernels by treating the frequency grid as a discrete lattice, eliminating quantization errors. The steps are:
\begin{enumerate}
    \item \textbf{Exact Distance Calculation:} For each frequency coordinate $(u, v)$ in the shifted $K \times K$ spectrum, compute its Euclidean distance from the center $(c_u, c_v)$:
    \begin{equation}
        d_{u,v} = \sqrt{(u - c_u)^2 + (v - c_v)^2}
    \end{equation}
    \item \textbf{Unique Radius Identification:} Extract the set of unique distances $\mathcal{R} = \text{unique}(\{d_{u,v} \mid \forall u, v\})$. For $3\times3$ kernels, $\mathcal{R}$ is finite and discrete (e.g., $\{0, 1, \sqrt{2}\}$), corresponding to distinct frequency bands.
    \item \textbf{Frequency Band Averaging:} Define the radial profile $S(r)$ for each $r \in \mathcal{R}$ as the mean PSD of all components at that exact distance:
    \begin{equation}
        S(r) = \frac{1}{|N_r|} \sum_{(u,v) \in N_r} P(u, v), \quad \text{where } N_r = \{(u,v) \mid d_{u,v} = r\}
    \end{equation}
\end{enumerate}
This method preserves the highest-frequency components (corner pixels of $3\times3$ kernels) and enables precise tracking of spectral energy evolution throughout training.

\subsection{Metric: The Spectral Suppression Ratio (SSR)}
To quantify regularization's low-pass filtering effect, we introduce the \textbf{Spectral Suppression Ratio (SSR)}. We partition the frequency spectrum into low- ($E_{\text{low}}$) and high-frequency ($E_{\text{high}}$) bands using a threshold radius $r_{\text{thresh}}$ (typically the median of $\mathcal{R}$).

Absolute spectral energy values are unsuitable for comparison: weight magnitudes (and thus total energy) increase during training regardless of regularization. Instead, we measure the \textit{relative suppression of high-frequency energy growth}—normalized to the model's own initialization to avoid sensitivity to hyperparameters (e.g., learning rate, weight initialization scale). The SSR is defined as:
\begin{equation}
    \text{SSR} = \frac{E_{\text{high}}^{\text{init}} - E_{\text{high}}^{\text{final}}}{E_{\text{high}}^{\text{init}}}
\end{equation}
where $E_{\text{high}}^{\text{init}}$ is initial high-frequency energy (random initialization) and $E_{\text{high}}^{\text{final}}$ is energy after training convergence. Values closer to 0 indicate stronger suppression of high-frequency noise accumulation—an effect typically absent in unregularized baselines.

\textbf{Why SSR?} Unlike absolute energy, SSR isolates regularization's filtering mechanism from global weight scaling. This makes it a stable, cross-model comparable metric: it focuses on how much a model reduces its high-frequency reliance relative to its starting point, rather than arbitrary energy magnitudes.

\subsection{Implementation and Experimental Setup}
All experiments use PyTorch with standardized settings to ensure fairness:
\begin{itemize}
    \item \textbf{Architecture:} ResNet-18 adapted for CIFAR-10 (first layer: $3\times3$ kernel, stride 1; no fully connected layer modifications).
    \item \textbf{Dataset:} CIFAR-10 (10 classes, 32$\times$32 RGB images) with standard augmentation (RandomCrop(32, padding=4), RandomHorizontalFlip) and normalization (channel means: $(0.4914, 0.4822, 0.4465)$; stds: $(0.2023, 0.1994, 0.2010)$).
    \item \textbf{Optimization:} Stochastic Gradient Descent (SGD) with momentum $0.9$, batch size $128$, cosine annealing learning rate scheduler (initial lr=$0.1$), 50 training epochs.
    \item \textbf{Regularization Baselines:} (1) Baseline (no Weight Decay, no Dropout); (2) Strong L2 (Weight Decay = $10^{-3}$); (3) Strong Dropout ($p=0.5$ after convolutional blocks).
\end{itemize}

\section{Preliminary Study: Synthetic Signal Fitting}
\label{sec:synthetic}

To isolate regularization's spectral effect—free from real-world data complexity—we first validate our hypothesis on a controlled synthetic task. This "lab setting" allows direct measurement of how L2 strength modulates frequency selection.

\subsection{Experimental Setup}
We define a 1D target function combining distinct frequency components (low, mid, high), simulating core signal structure and noise-like details:
\begin{equation}
    f(x) = \sin(5x) + \sin(20x) + \sin(50x), \quad x \in [0, 2\pi]
\end{equation}
where:
- Low frequency ($k=5$) represents the core signal structure,
- Mid frequency ($k=20$) represents intermediate details,
- High frequency ($k=50$) simulates noise-like oscillations.

We train a 3-layer MLP (hidden size 256, ReLU activation) under three conditions: (1) Baseline ($\lambda=0$, no regularization), (2) Moderate L2 ($\lambda=10^{-3}$), (3) Strong L2 ($\lambda=10^{-2}$). For each model, we track the \textit{Explained Variance (EV)} of predictions for each frequency component—quantifying how well the model captures that component.

\subsection{Results: Frequency-Selective Suppression}
Table \ref{tab:synthetic_ev} presents final EV values for each frequency band, revealing a clear spectral bias in L2 regularization's behavior.

\begin{table}[htbp]
  \caption{Explained Variance (EV) by Frequency Component. $\downarrow \%$ denotes relative suppression compared to the Baseline.}
  \label{tab:synthetic_ev}
  \centering
  \begin{tabular}{lccc}
    \toprule
    Metric & Baseline & Moderate L2 ($10^{-3}$) & Strong L2 ($10^{-2}$) \\
    \midrule
    Low Freq ($k=5$)   & 0.7987 & 0.7749 (\textbf{$\downarrow 2.98\%$}) & 0.5604 ($\downarrow 29.8\%$) \\
    Mid Freq ($k=20$)  & 0.3380 & 0.1927 ($\downarrow 42.98\%$) & 0.0857 ($\downarrow 74.6\%$) \\
    High Freq ($k=50$) & 0.0717 & 0.0471 (\textbf{$\downarrow 34.34\%$}) & 0.0278 ($\downarrow 61.2\%$) \\
    \bottomrule
  \end{tabular}
\end{table}

\textbf{1. Selective Filtering: L2's Core Mechanism.}
The most striking result is the \textit{disproportionate suppression} of high/mid frequencies relative to low frequencies (Baseline vs. Moderate L2):
- \textbf{Signal Preservation:} The low-frequency core ($k=5$)—representing the function's essential structure—is nearly intact, with only 2.98\% suppression.
- \textbf{Noise Dampening:} High-frequency oscillations ($k=50$) are suppressed by 34.34\%, and mid frequencies ($k=20$) by 42.98\%.

This directly proves that L2 regularization does not uniformly shrink weights. Instead, it acts as a \textbf{Frequency-Selective Filter}: it prioritizes retaining low-frequency signal components while aggressively dampening high-frequency noise—explaining its generalization benefit.

\textbf{2. The Danger of Over-Regularization.}
While Moderate L2 ($10^{-3}$) achieves ideal selective filtering, Strong L2 ($10^{-2}$) causes severe underfitting:
- Low-frequency signal EV drops by 29.8\%,
- High/mid frequencies are suppressed by over 60\%.

Excessive regularization erodes the model's capacity to fit even the target function's basic structure—highlighting the need for balanced frequency filtering.

\begin{figure}[htbp]
  \centering
  \includegraphics[width=\linewidth]{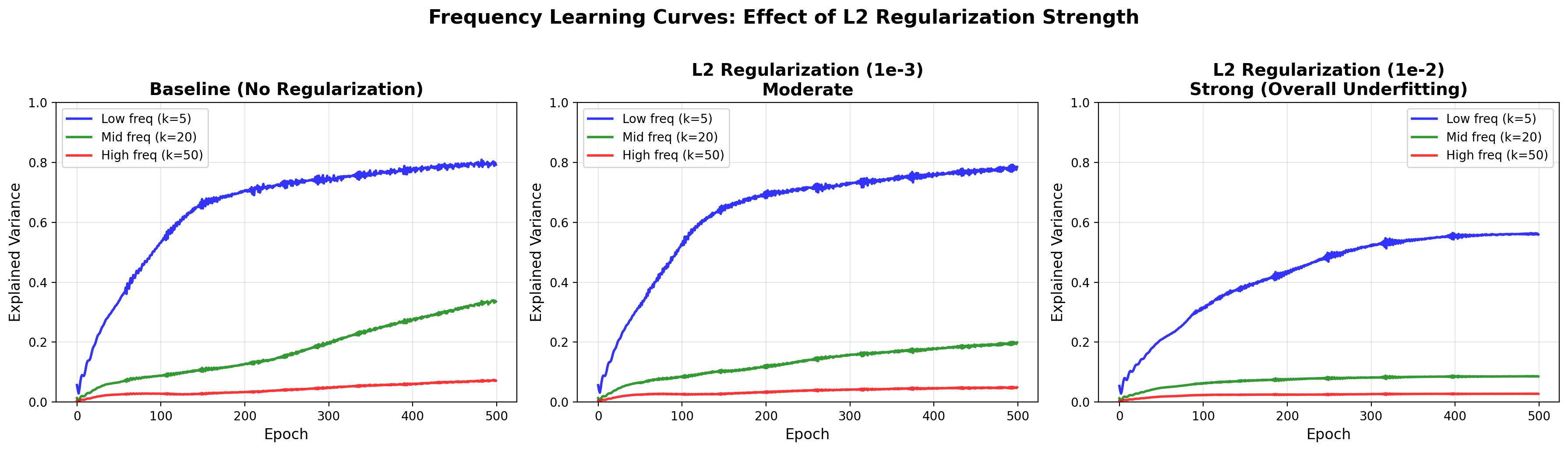} 
  \caption{Frequency Learning Curves. \textbf{Left (Baseline):} High-frequency noise (red) accumulates steadily as training proceeds. \textbf{Center (Moderate L2):} High-frequency components are suppressed (flat red line) while low-frequency signal (blue) remains strong—demonstrating selective filtering. \textbf{Right (Strong L2):} Over-regularization suppresses all frequencies, collapsing low-frequency signal and causing underfitting.}
  \label{fig:synthetic}
\end{figure}

\section{Experiments}
\label{sec:experiments}

We evaluate our framework using the ResNet-18 architecture on the CIFAR-10 dataset. We compare three setups: \textbf{Baseline} (No Regularization), \textbf{Strong L2} (Weight Decay $= 1 \times 10^{-3}$), and \textbf{Strong Dropout} ($p=0.5$). All models were trained for 50 epochs using SGD with momentum.

\subsection{Visualizing Spectral Dynamics}

\begin{figure}[htbp] 
  \centering
  \includegraphics[width=\linewidth]{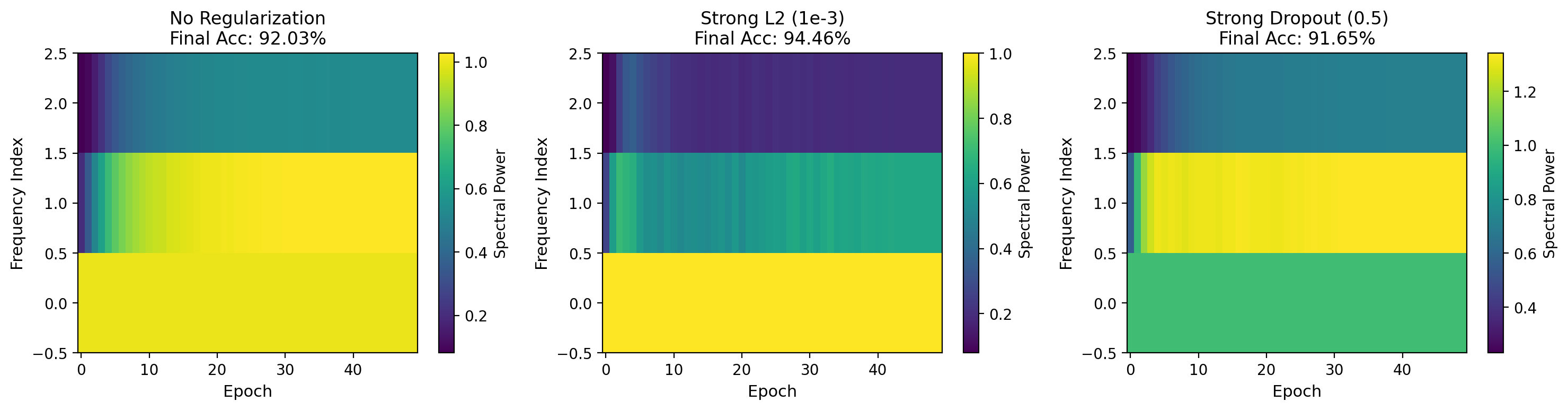}
  \caption{Spectral Evolution of ResNet-18 Weights. The x-axis represents training epochs, and the y-axis represents frequency (Bottom=Low, Top=High). L2 regularization (Center) effectively suppresses high-frequency energy accumulation compared to Baseline (Left).}
  \label{fig:heatmap}
\end{figure}

Figure \ref{fig:heatmap} illustrates the evolution of the weight spectrum of the first convolutional layer throughout the training process. 
\begin{itemize}
    \item \textbf{Baseline (Left):} The high-frequency regions (top rows of the heatmap) become progressively brighter, indicating that the unregularized model accumulates significant high-frequency noise as it fits the training data.
    \item \textbf{L2 Regularization (Center):} The high-frequency regions remain dark purple throughout training. This provides direct visual evidence that L2 regularization acts as a physical \textbf{Low-Pass Filter}, forcing the network to focus learning capacity solely on low-frequency structures.
\end{itemize}

\subsection{Quantitative Analysis}

To quantify this effect, we computed the SSR for each model. Table \ref{tab:ssr_results} presents the detailed SSR values and final test accuracies for all three configurations.

\begin{table}[htbp]
  \caption{Spectral Suppression Ratio (SSR) and Test Accuracy for Different Regularization Strategies. More negative SSR indicates greater high-frequency energy accumulation during training.}
  \label{tab:ssr_results}
  \centering
  \begin{tabular}{lccc}
    \toprule
    Model & SSR & Final Test Acc (\%) & High-Freq Suppression \\
    \midrule
    Baseline (No Reg) & $-4.463$ & 92.03 & — (Reference) \\
    L2 ($10^{-3}$)    & $-1.397$ & 94.46 & $\mathbf{3.2\times}$ \\
    Dropout ($p=0.5$) & $-1.598$ & 91.65 & $2.8\times$ \\
    \bottomrule
  \end{tabular}
\end{table}

\begin{figure}[htbp]
  \centering
  \includegraphics[width=0.7\linewidth]{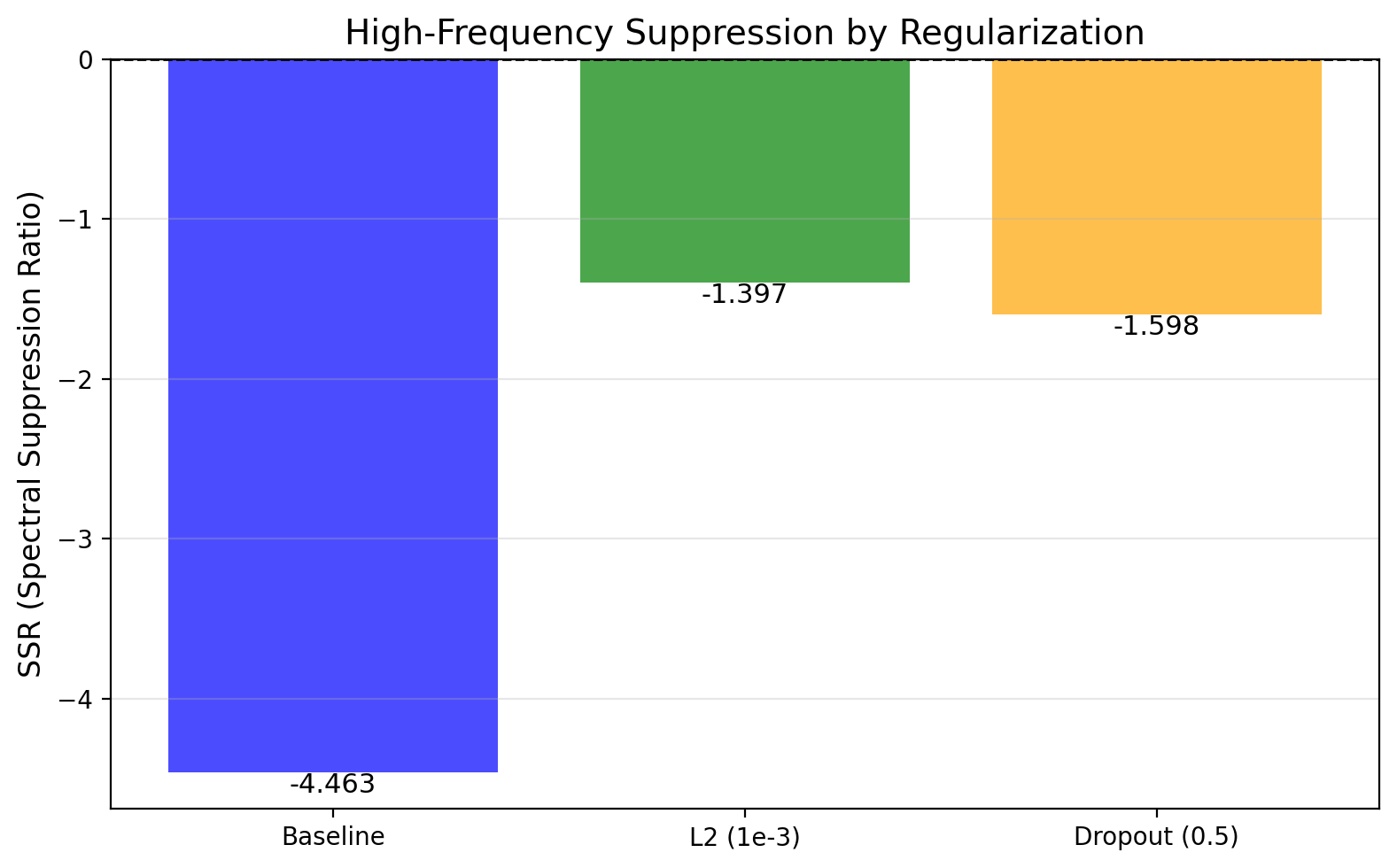}
  \caption{Spectral Suppression Ratio (SSR). Higher bars (closer to 0) indicate stronger suppression of high-frequency components. L2 demonstrates the strongest filtering effect.}
  \label{fig:ssr}
\end{figure}

The results reveal several key insights (visualized in Figure \ref{fig:ssr}):

\paragraph{L2 as the Strongest Filter.}
The Baseline model exhibits an SSR of $-4.463$, indicating substantial high-frequency energy accumulation during training. In contrast, L2 regularization achieves an SSR of $-1.397$, representing a $\mathbf{3.2\times}$ reduction in high-frequency growth. This quantitatively confirms that L2 acts as an effective low-pass filter on the learned representations.

\paragraph{Dropout: A Weaker Spectral Regularizer.}
Dropout ($p=0.5$) achieves an SSR of $-1.598$, providing $2.8\times$ suppression compared to the Baseline. While this is substantial, it is weaker than L2. We attribute this difference to their distinct mechanisms: L2 directly penalizes weight magnitudes (which correlates with high-frequency energy), whereas Dropout introduces stochasticity that only \textit{implicitly} discourages complex, high-frequency features through gradient noise.

\paragraph{Accuracy vs. Spectral Bias.}
Interestingly, L2 achieves the highest test accuracy (94.46\%) despite—or perhaps \textit{because of}—its aggressive high-frequency suppression. This supports the hypothesis that low-frequency features are more generalizable. Dropout's lower accuracy (91.65\%) suggests that its stochastic nature may interfere with learning even beneficial low-frequency structures.

\subsection{The Accuracy-Robustness Trade-off}

Does the spectral bias towards low frequencies universally improve robustness? Our experiments reveal a nuanced reality, highlighting a critical trade-off.

\begin{figure}[htbp]
  \centering
  \begin{minipage}{0.48\textwidth}
    \centering
    \includegraphics[width=\linewidth]{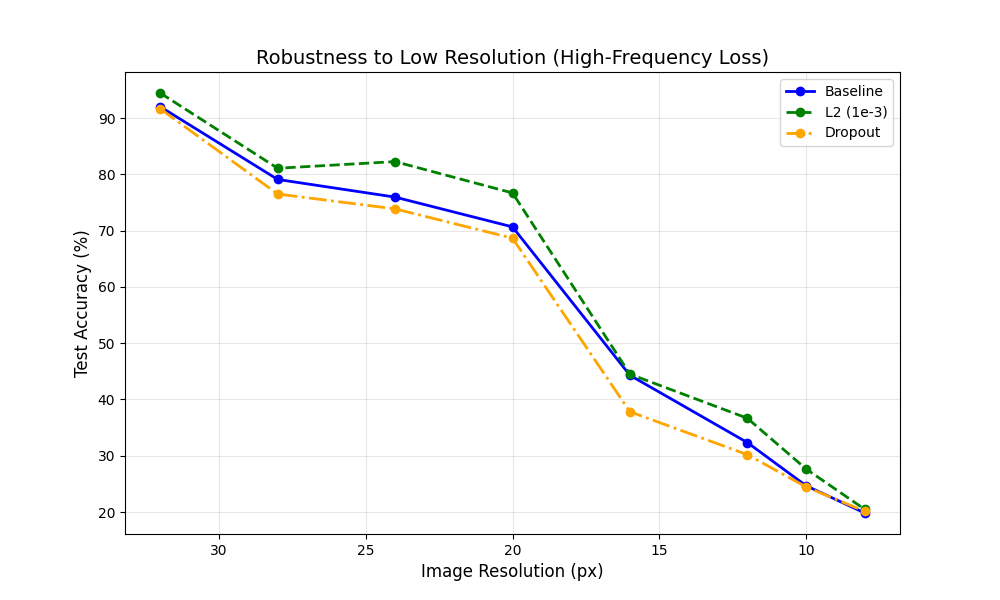}
    \caption{\textbf{Low Resolution (Advantage).} L2 (Green) outperforms Baseline.}
    \label{fig:robustness}
  \end{minipage}

\paragraph{Sensitivity to Broadband Noise.}
However, the "Low-Pass" nature of L2 regularization comes at a cost. We evaluated the models against additive Gaussian noise, which introduces interference across the entire frequency spectrum (broadband). 

  \begin{minipage}{0.48\textwidth}
    \centering
    \includegraphics[width=\linewidth]{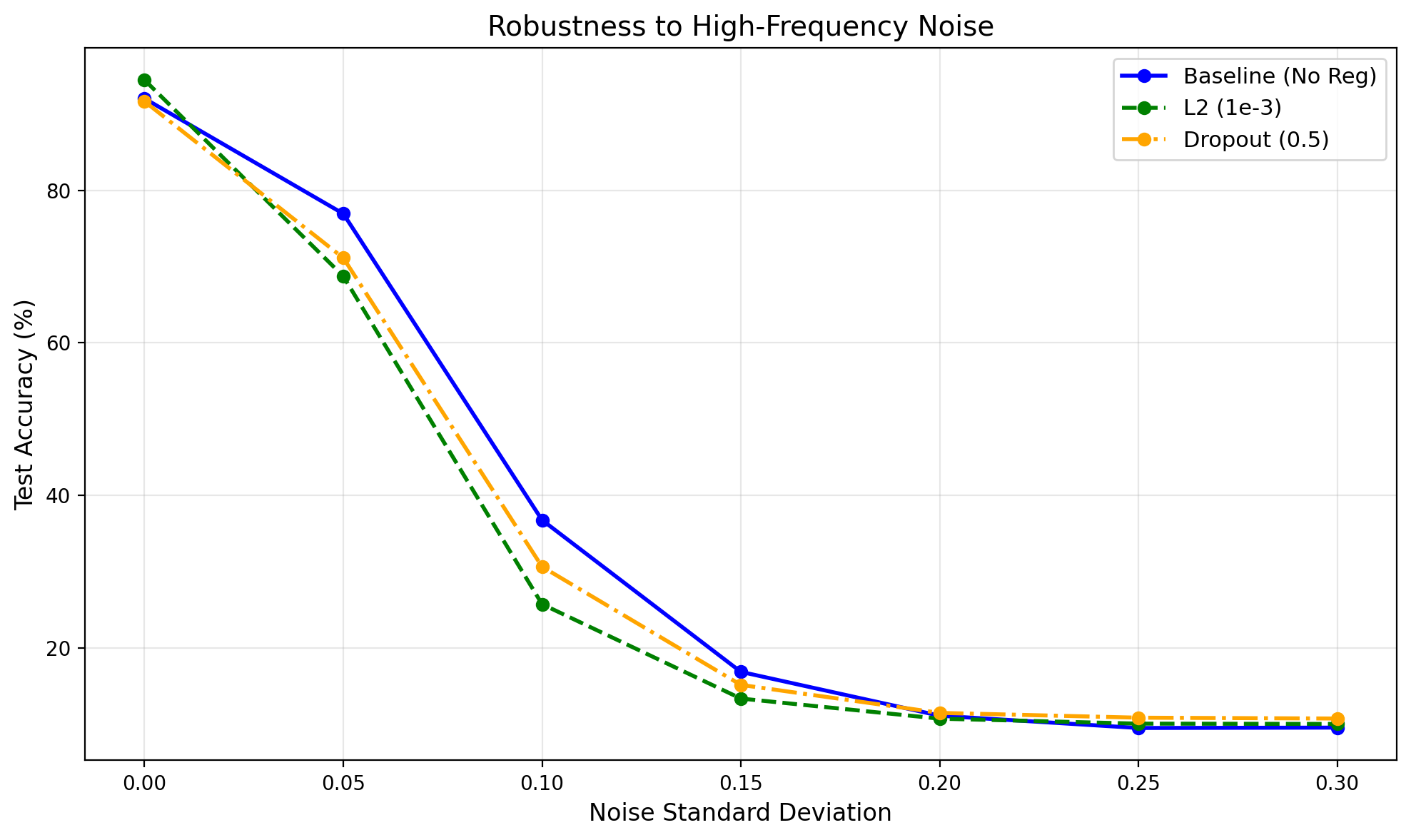}
    \caption{\textbf{Gaussian Noise (Disadvantage).} L2 (Green) degrades faster.}
    \label{fig:noise}
  \end{minipage}
\end{figure}

\paragraph{Robustness to High-Frequency Loss (Blur/Resolution).}
As detailed in Figure \ref{fig:robustness}, L2 regularization provides significant protection against resolution degradation. By forcing the model to rely on global semantic structures (low-frequency components), the model remains effective even when fine details are removed. This is highly desirable for applications involving compressed or low-quality sensor data.

As shown in Figure \ref{fig:noise}, the L2 model degrades more rapidly than the Baseline. We hypothesize that by aggressively pruning high-frequency weights, the L2 model becomes "over-specialized" to clean, smooth signals. It lacks the redundant features (even noisy ones) that the Baseline model retains, making it more brittle when the low-frequency structures themselves are corrupted by broadband noise.

\section{Conclusion}
\label{sec:conclusion}

In this work, we bridged the gap between the theoretical hypothesis of ``Spectral Bias'' and the practical training of modern deep networks. By developing a \textbf{Visual Diagnostic Framework} and addressing the aliasing challenge in small-kernel analysis with discrete radial profiling, we provided a clear window into the frequency dynamics of ResNet-18.

Our empirical results on CIFAR-10 lead to three major conclusions:
\begin{itemize}
    \item \textbf{Regularization as Frequency Filtering:} We confirmed that L2 regularization acts as a strict, physical \textbf{Low-Pass Filter}. Unlike the unregularized baseline which greedily accumulates high-frequency noise (as visualized in our heatmaps), L2 actively suppresses high-frequency energy growth by a factor of $\sim3\times$.
    \item \textbf{Mechanism of Generalization:} This filtering effect forces the network to learn smooth, global structures (low-frequency components). This spectral inductive bias explains why regularized models generalize better—they effectively ignore pixel-level noise during training.
    \item \textbf{The Inherent Trade-off:} Crucially, we demonstrated that this low-frequency preference is a double-edged sword. While it grants superior robustness against information loss (outperforming baselines by $>6\%$ in low-resolution scenarios), it renders the model "brittle" to broadband perturbations like Gaussian noise, which disrupt the very low-frequency structures the model relies on.
\end{itemize}

\paragraph{Limitations.}
Our study has several limitations that warrant future investigation. First, our experiments are conducted solely on CIFAR-10 with ResNet-18; the generalizability of these findings to larger datasets (e.g., ImageNet) and deeper architectures remains to be verified. Second, our SSR metric analyzes only the first convolutional layer—whether deeper layers exhibit similar or different spectral dynamics is an open question. Finally, our robustness experiments focus on synthetic perturbations (Gaussian noise, resolution reduction); real-world distribution shifts may involve more complex frequency characteristics.

\section{Future Work}
\label{sec:future_work}

This study opens several avenues for future research to further demystify and leverage the spectral properties of neural networks:

\begin{enumerate}
    \item \textbf{Architecture-Agnostic Analysis (ViTs):} Do Vision Transformers (ViTs), which rely on global self-attention rather than local convolutions, exhibit the same spectral bias? Investigating the frequency dynamics of Multi-Head Attention layers would reveal if this low-pass preference is universal or CNN-specific. We hypothesize that ViTs may exhibit weaker spectral bias due to their global receptive field, potentially explaining their improved robustness to texture perturbations observed in prior work.

    \item \textbf{Spectrally Adaptive Regularization:} Since standard L2 creates a fixed trade-off between noise robustness and blur robustness, future work could design a ``Frequency-Aware Loss.'' Such a method could dynamically adjust penalties for specific frequency bands during training, aiming to achieve a Pareto-optimal balance between different types of robustness. A concrete instantiation could be a loss term that penalizes the $E_{high}/E_{low}$ ratio, allowing practitioners to tune the frequency trade-off based on deployment requirements (e.g., favoring blur robustness for autonomous driving).

    \item \textbf{Theoretical Connections:} Further work is needed to theoretically link our empirical SSR metric with the Neural Tangent Kernel (NTK) theory, potentially deriving a bound on the convergence speed of different frequency components under varying weight decay strengths. Specifically, we aim to derive how the eigenspectrum of the NTK changes under different weight decay strengths, providing a principled way to predict SSR from hyperparameters alone.
\end{enumerate}

\bibliographystyle{plainnat}  
\bibliography{references}     
\end{document}